# 2020 NASA RASCA-AL Special Edition:
## Moon to Mars Ice & Prospecting Challenge

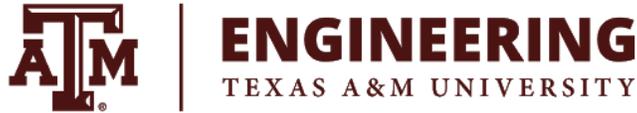

**Technical Paper**

# DREAMS: Drilling and Extraction Automated System

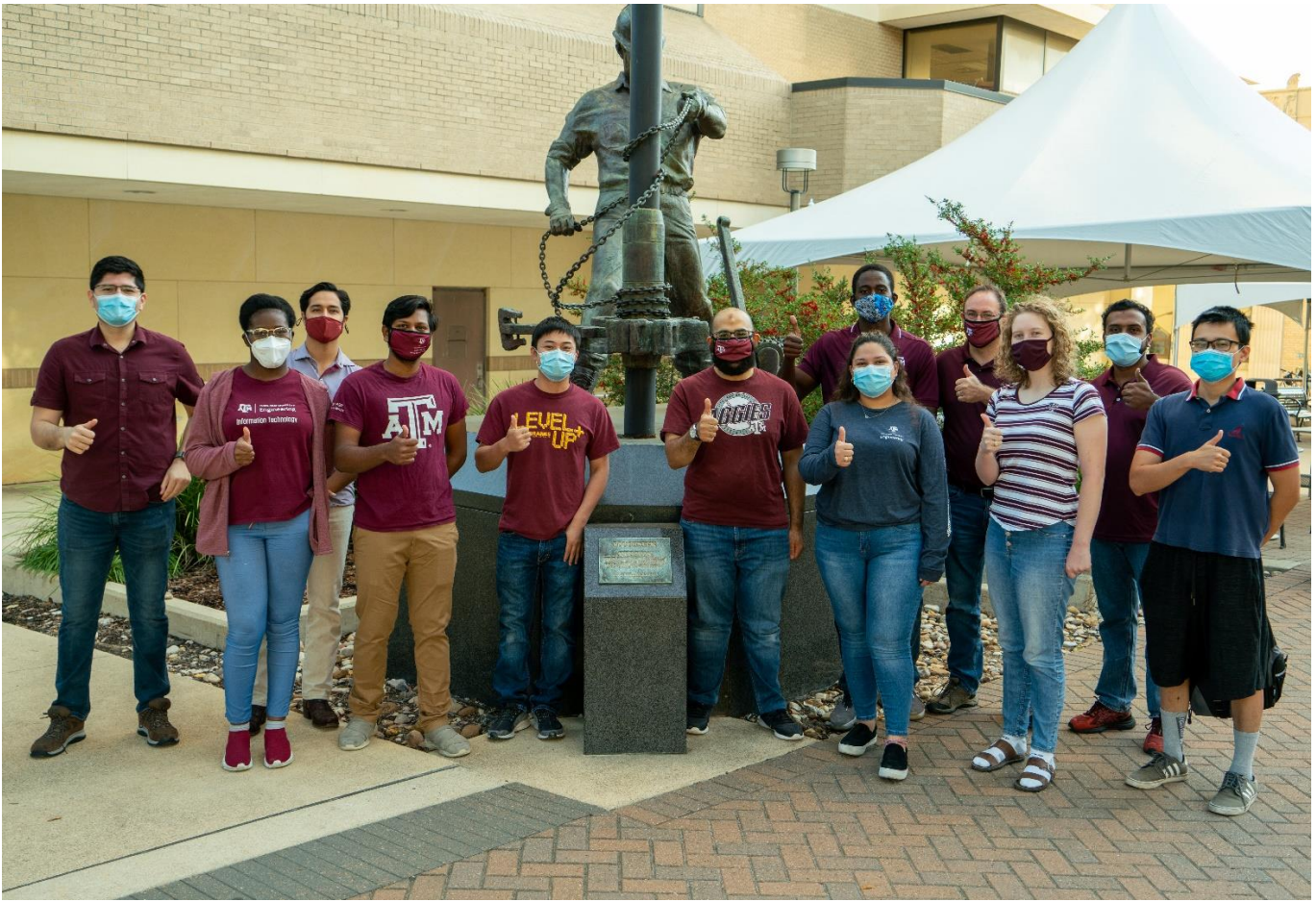

### Team Members
Mohamed Khaled, Srivignesh Srinivasan, Alkassoum Toure, Muhao Chen, Emily Kincaid, Thomas Lopaz, Luis Rodriguez, Jessica Ezemba, Ayodeji Adeniran, Teresa Valdez, Uthej Vattipalli, Le linh, and Ahmed Madi

### Faculty Advisors
Dr. Eduardo Gildin, Dr. Robert Skelton, Dr. Sam Noynaert, and Dr. George Moridis

| Team Members | Major | Academic Degree |
|---|---|---|
| Ahmed Madi | ME | 2021 |
| Ayodeji Adeniran | ME | PhD, 2023 |
| Alkassoum Toure | PETE | Masters, 2022 |
| Emily Kincaid | ME | 2021 |
| Jessica Ezemba | ME | 2021 |
| Le Linh | PETE | Masters, 2020 |
| Luis Rodriguez | PETE | Masters, 2022 |
| Mohamed Khaled (Team Lead) | PETE | PhD, 2021 |
| Muhao Chen | AE | PhD, 2021 |
| Srivignesh Srinivasan | ME | Masters, 2020 |
| Teresa Valdez | CE | 2021 |
| Thomas Lopez | IDE | PhD, 2022 |
| Uthej Vattipalli | CE | Masters, 2020 |

*Summary*

Drilling and Extraction Automated System (DREAMS) is a fully automated prototype-drilling rig that can drill, extract water and assess subsurface density profiles from simulated lunar and Martian subsurface ice. DREAMS system is developed by Texas A&M drilling automation team and composed of four main components: 1- tensegrity rig structure, 2- drilling system, 3- water extracting and heating system, and 4- electronic hardware, controls and machine algorithm. The vertical and rotational movements are controlled by using Acme rod, stepper and rotary motor. DREAMS is a unique system and different from other systems presented before in NASA Rascal-Al competition, because 1- It uses tensegrity structure concept to decrease the system weight, improve mobility and easier installation in space. 2- It cuts rock layers by using short bit length connected to drill pipes. This drilling methodology is expected to drill hundreds and thousands of meters below moon and Martian surfaces without any anticipated problems (not only 1 m.). 3- Drilling, heating and extraction systems are integrated in one system that can work simultaneously or individually to save time and cost.

*Mounting system*

For this year's competition, a wooden plate was manufactured to attach the tensegrity structure and mount easily to the mounting box supplied by the NASA competition. The mounting piece was designed so that a worm box could be situated under the ACME motor rod. On top of that, it helps further increase the stability because the rectangle base will be bolted to fiber foam support that is fixed to the welded "NASA table." The worm box was installed because the stepper motor that oversees the vertical motion of the drilling rig did not fit in the previous design. The tensegrity structure will be bolted onto the fiber foam support because the tensed cables will be situated onto this board as well. This is crucial for the tensegrity structure because it will increase the strength and stability of the structure to drill efficiently. The fiber foam support is then bolted onto the steel wooden to further increase robustness of the structure and ensure that no slipping will occur from the vibrations of drilling. The wooden table is then placed on top of the NASA mounting box to be able to drill into the rock and ice specimen securely. This shape was chosen to maximize the robustness of the table and guarantee that the drilling rig and the important components would be supported adequately.

*System description*

**Tensegrity Rig Structure**

Skelton, et al., 2009 demonstrated that T-Bar and D-Bar tensegrity systems require much less mass than a single continuum bar under compression. Tensegrity solutions provide the minimal mass for all of the fundamental structural loading conditions in engineering mechanics. Tensegrity is a network of bars and strings, where bars only take compression and strings take tension. All structural members are axially loaded, hence the structural efficiency in strength to mass is very high. The team's main reason for using a tensegrity structure is to reduce the weight of the traditional drilling rig without having to sacrifice the integrity of the structure. The structure is composed of six unique components that contribute to the success of the structure as shown in Fig. 1 and in additional photos located in Appendix A. The two rings highlighted in purple are responsible for supporting the drilling rig's railing system and structure. The bottom ring will be bolted to the mounting system, connect the ring joints (green), have a through-hole for the acme motor rod (blue), and four threaded holes for the stability rods (orange). The top ring has all the same tasks as the bottom ring except that all holes will be through holes and it won't have to be bolted. The ring joint (bright green) will connect all the tensegrity rings through the access hub. The ring joint would be attached using a pin going through the holes of the access hub and the ring joint. The integration of these pieces allows for the connecting rods (grey) of the structure to be attached. The class 2 joint (yellow) was designed to be able to integrate the connection between the top and bottom rings with the middle ring and to assemble with the C-joint (red). A connecting rod is inserted into the insert of the class 2 joint to connect to either the middle, bottom, or top. A ball bearing will be situated inside the circle cutout so that it can rotate freely when assembled with another class 2 joint. The C joint was designed to be able to crimp three steel cables independently such that the each cable can be set in tension at a 45-degree angle towards the base or other C-joints to help improve the stability of the structure greatly. The middle ring was designed to connect the top and bottom rings, allow passage of the mobile ring, and allow a connection for the class 1 joint. The ring has an inner diameter of 6" to allow the railing system's mobile ring to move up and down freely. The access hubs here were increased in size to allow for the connections of two ring joints and one class 1 joint. The figure below illustrates the middle ring. Lastly, the class 1 joint was designed to crimp three steel cables in tension that are connected to the base of the structure and to both c joints above and below. Moreover, there is an insert for a connecting rod to be inserted to be merged with the middle ring, as described earlier. The figure below illustrates the class 1 joint.

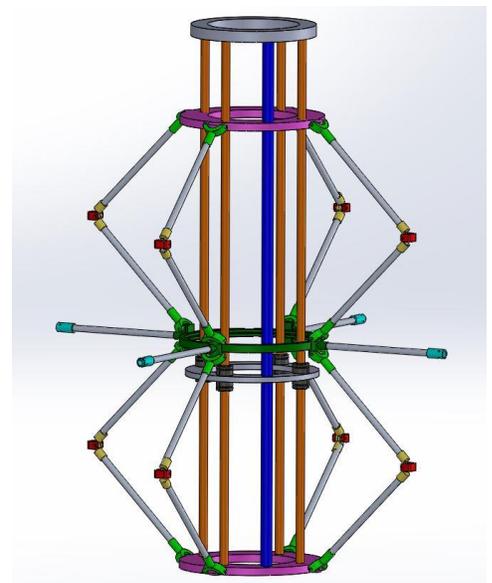

Fig. 1. Tensegrity Network

Beside the main six components of tensegrity structure, the structure also comprised of drillstring, acme motor, gearbox, gearbox mount, hydraulic swivel, drillstring casing, heating element, etc. as shown in Fig. 2. A gearbox mount was designed and manufactured to secure the gearbox to the tensegrity rig assembly. This piece was manufactured to help stabilize the motor and drillstring assembly inside the tensegrity rig. Furthermore, it helps keep the system from becoming unbalanced during the drilling operation. The drillstring casing is manufactured from wrapping heating element around the casing, the system was coated with Sauereisen Ceramic Cement to make it electrically insulated and be able to withstand extreme temperatures produced to melt the ice.

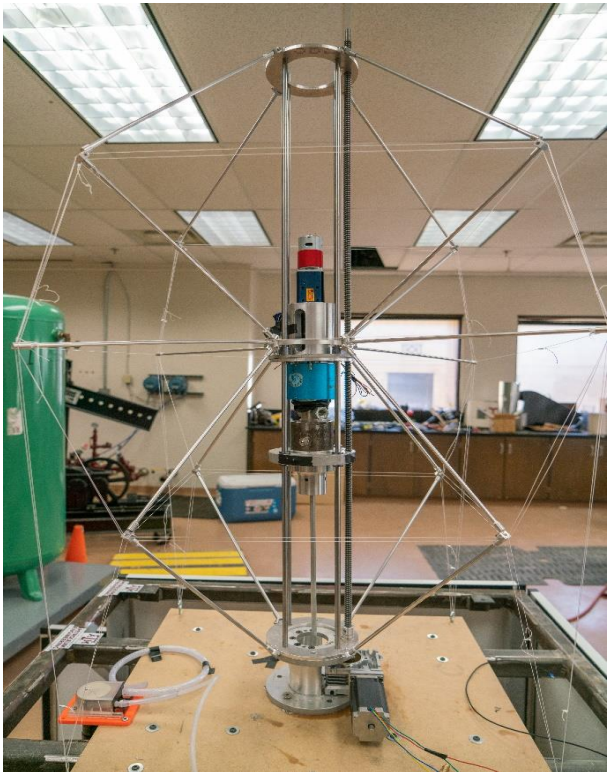
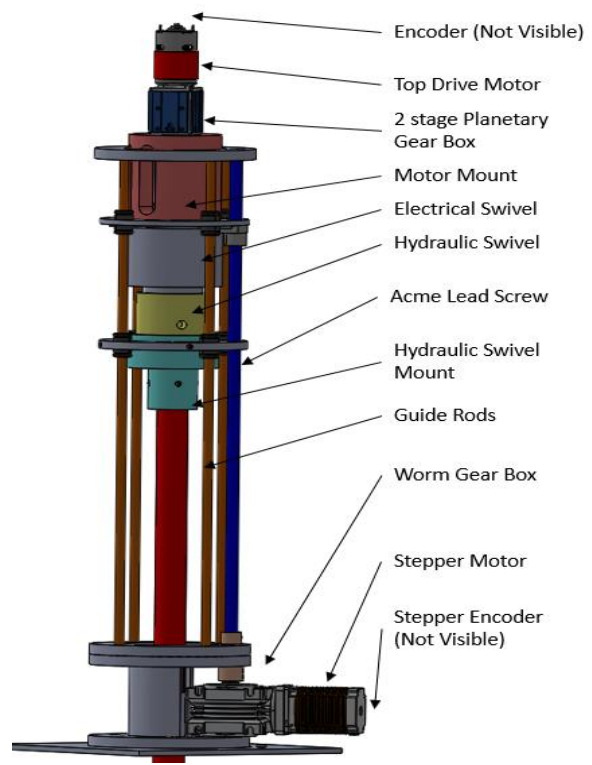

**Fig. 2**. Tensegrity structure and mounting system

**System Excavation Operations**
The DREAMS structure will be placed on top of the sample rock and ice and the operator will command the drilling rig to begin drilling operations. The open loop control algorithms will control the drilling automations through different rock samples and ice. The rotary motor at the top of the structure will rotate the drill string, bottom hole assembly (BHA), and the drill bit, while the stepper motor and ACME rod will lower the drill bit until it contacts the overburden layers. A connector piece known as the "merging piece" was manufactured to attach the casing to the motor-drill string assembly. The heating element is going to be wrapped around the casing to transfer the necessary heat for melting the ice. A drill string is inside the casing and attached to the BHA cap. The drill string has an outer diameter of 0.375" in and a wall thickness of 0.035" in. The BHA connects the drill string to the drill bit, and is also made of aluminum. DREAMS used a Baker Hughes micro bit with 1.5" in OD and 1.48" in makeup length. The cap is connected to the bit and string with a 3/8-18 NPT standard sized thread on the outside and with a 3/4-32 NPT thread on the inside of the cap that is connected to the BHA sensor capsule. The capsule has a 0.9" in diameter and functions primarily to mount sensors, protect sensors from melted water, aid sensors in performing accurately, and collecting, retrieving, and sending data to the top of the rig for further post processing.
Within the BHA sensor capsule are two holes for the melted ice water to flow from the drill bit to the drill string and back, allowing for the sensor to take appropriate readings. The module holding the sensor has a hole on one end attached to the drill string and facing away from the drill bit that allows the wires to connect to the hydraulic swivel. This system uses all electric connections. Sealant is put in the cavity surrounding the sensor so no water touches the sensors. The drill string, BHA casing, and drill bit assembly are shown in Fig. 3 respectively.

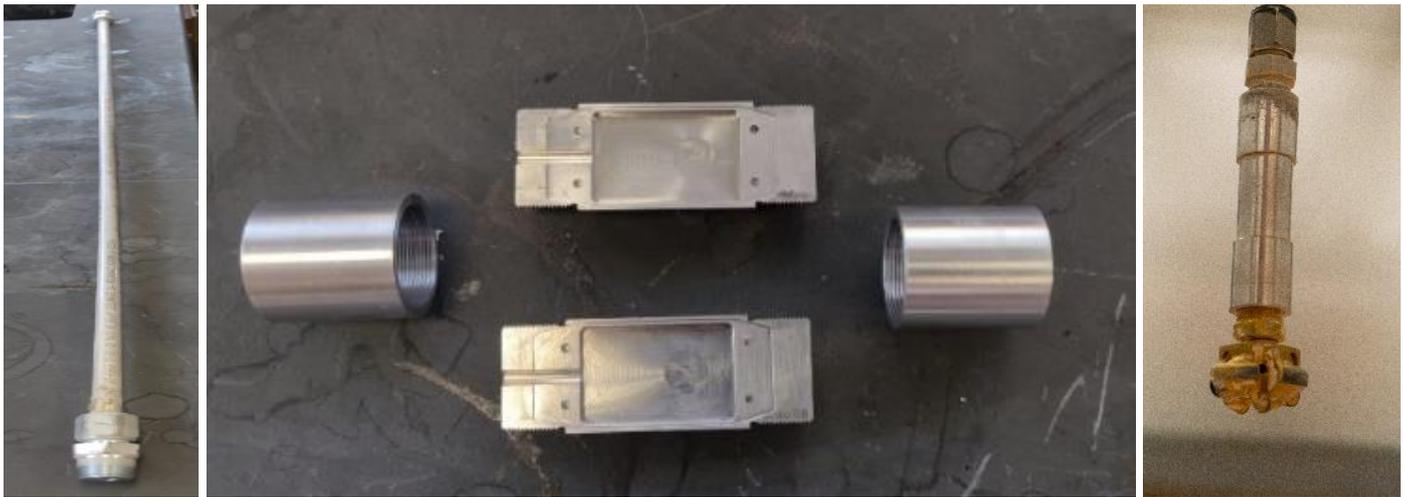

*Fig. 3.* Drill String, BHA Casing, and Drill Bit Assembly

The drilling and extraction operation will consist of two main steps. The first step is to drill the rock and ice with the drill bit. While drilling, air will be pumped through the hydraulic swivel, drill string, and BHA and drill bit to clear the hole of unwanted solid particles. After this step is completed, the heating element will kick in. The heating element is composed of high resistance nichrome wires (Fig. 4) that are wrapped evenly along the outside of the drill casing almost 18" in from the bottom. An anticipated 200-600 W is supplied by the nichrome wires and the surrounding ice is heated through radiation and conduction from the wires. As the ice melts and water surrounds the wires the heat transmission changes from radiation to purely conduction. The casing may get heated from the wires as an indirect consequence of the heating process, but that won't cause any issues as the casing is coated in Sauereisen Ceramic Cement to protect vulnerable and fragile elements from excessive heat and possible electricity.

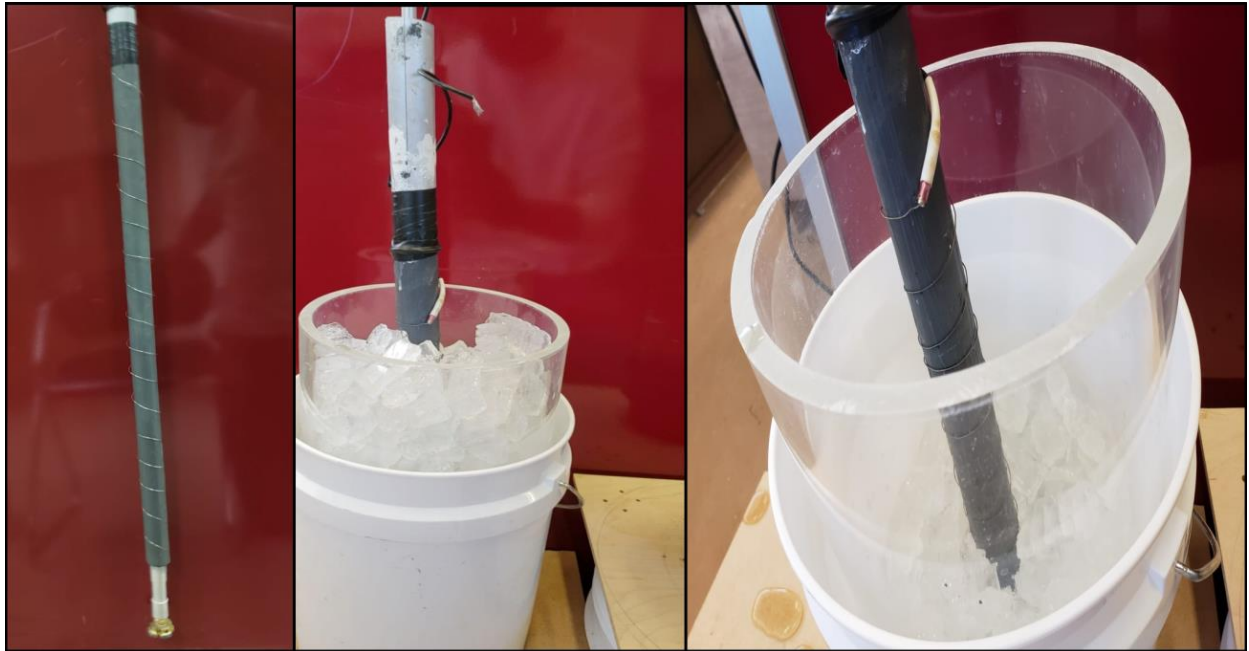

**Fig. 4.** 3 in 1 Drilling/Heating/Production unit (left) – Ice melting (center) – 1hr Final result (right)

When the heating process is completed, the extraction of the water will start. The three-way control valve will switch from the air pump to the water pump. The melted water running through the sides of the BHA chain acts indirectly as a cooling fluid to prevent any unwanted heating of the sensor and wires. As a part of future modification plans, the heating and water extraction steps can be alternated in the event that the ice is deeper in the sample. This whole process will be repeated until time runs out or the team is satisfied with the water recovery. At present system design and with a power supply of 200W, we were able to obtain a melting capacity of 1570cc/hr. We are anticipating a better melting rate which would most likely double or triple with an increase in power supply to 400 or 600W. As water reaches the top, it is passed through 2 layers of filtering material that is detailed in the Water Extraction, Filtration, and Collection section of this report. The system described can be viewed as shown in Fig. 5.

**Water Extraction, Filtration and Collection**
After the ice is melted, it is pumped to the surface using a peristaltic pump. The produced water must travel through a mesh filter before entering the pump to prevent sizable debris from entering and damaging the pump. The water then continues to make its way to the filtration unit. During development of the filtration unit design, the team considered a conical versus a cylindrical design. After trialing both designs, the conical design was selected, and the prototype was built. The final filtration unit design has a higher cross-sectional area (6") and a broader conical opening diameter (3") than previous designs. Factors that affected the dimensional configurations included, but were not limited to, available space on the rig, overall water extraction and filtration rates, and parts availability.

After running tests with some initial ideas, the team noticed that there was always a trade-off between filtration rate and filtration quality. For better filtration quality, there had to be sacrifices in the filtration rate. The selected filtration design is able to provide an intermediate rate of water filtration and better water quality.

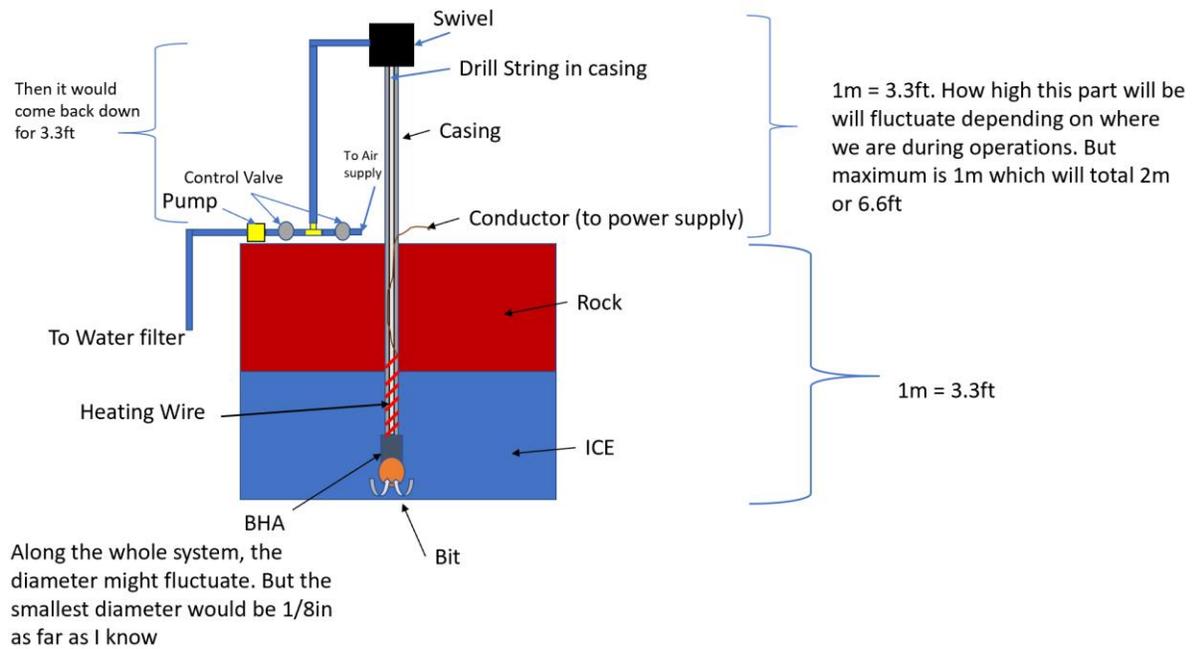

*Fig. 5.* Heating, Extraction and filtration systems schematics

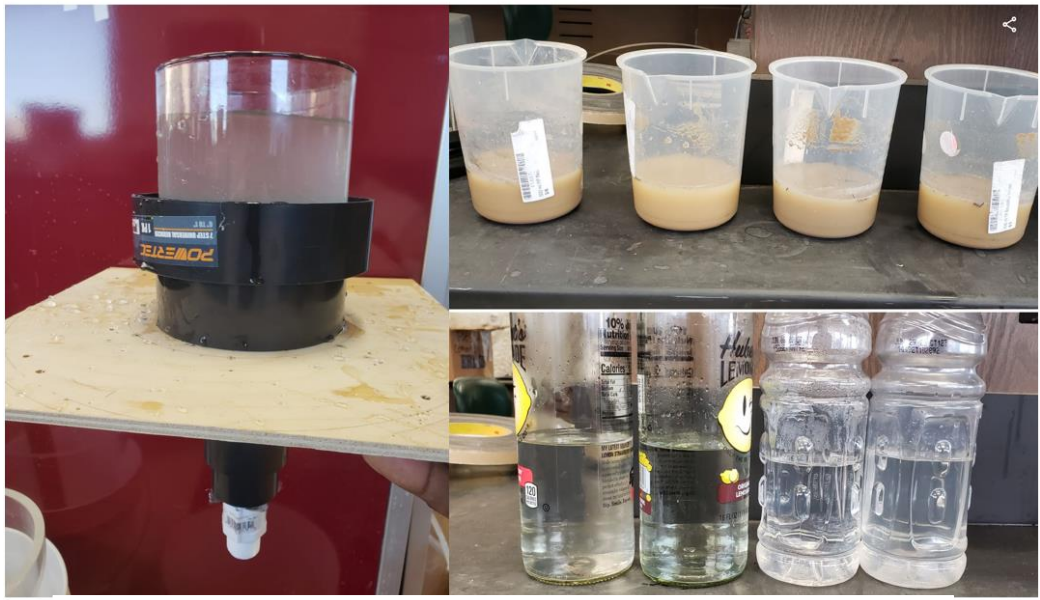

**Fig. 6.** Filtration Unit Schematics

Selecting the right materials to construct the filtration layer required many tests. The team considered 100 mesh olivine sand, 20 mesh silica sand, gravel, activated carbon, grade 1 filter paper, grade 4 filter paper, filtration foam and cotton balls. Experiments were conducted by varying the filter paper grade, sand grade, gravel presence, and activated carbon presence, leading to a total of around 40 experiments. The current filtration layer consists of a 2-to-1 ratio of 100 mesh olivine sand to 20 mesh silica sand. A wild card in the unit is a cotton ball fitted at the bottom. It was originally intended to act as soft support for the layers of sand above, but through experimentation we discovered that water quality was also improved with its presence. Ultimately, it was discovered through experimentation that a higher flow rate would be yielded with an enlarged funnel for the filtration layers. As such, with current design we yielded a filtration flow rate of 750cc/hr along with an improved water quality, as shown in Figure 6.

**System Methodology utilized for prospecting for a digital core**

A novel rock formation identification technique using machine learning models has been designed for this project. Tests had been conducted only on consolidated sandstone, fluffy sand, and concrete. Data contains rotary motor raw voltages and accelerometers recordings that the rig uses to observe downhole vibration data. We have one output variable for this problem that is formation type. The algorithms was designed to classify any given formation into one of the 5 classes - Sand (1), Clay (2), Stone (3), Concrete (4) and Ice (5). The classification analysis was divided into two steps: the best predictors/features extraction and classification model building. The models are supposed to be trained using multiple classification algorithms namely logistic regression, linear discriminant analysis (LDA), Support Vector Machines (SVM), Random Forest (RF) and Artificial Neural Networks (ANN). It is worth to mention that although we built our machine learning codes using different algorithms, we were unable to validate these codes due to the limited data recorded while experimental drilling tests.

**Control and communication system**

Electronic components of DREAMSDREAMS will be categorized into four sections: Sensors, Motors, Data Acquisition, and Power Requirements.

1-Sensors: The sensors in DREAMS are used for gaining information about the rock surface for further automation of the process, ensuring the reliability and efficiency of the system, and controlling the safety parameters that the system has to meet. The sensors in the system are detailed as follow. a- An Inertial Measurement Unit (IMU): An IMU is a sensor that measures the exerted force, angular rate, and orientation of a body with accelerometers. b- Temperature Sensor: The temperature of DREAMS is crucial to the viability of the sensors and the information to process the data. DREAMS must not exceed the maximum serviceable temperature of all sensors and Data Acquisition (DAQ) to ensure accurate and reliable results.

2- Motors: DREAMS is equipped with two main motors; the drilling stepper drive motor and the .top drive motor. a- Top Drive Motor: is located above the gearbox and controls the torque of the drill string. The drill string will need higher power input to breach varying terrains. Because of this, the top drive motor is a light weight, ball bearing supported, air-cooled, faster and a more powerful motor. This motor is rated up to 363 Watts and easily connects with the existing gearbox in the DREAMS design. This motor will also be installed with spacers to ensure sufficient cooling. b- Drilling Drive Motor: is a stepper motor that is used to control the up-down motion (Z motion) that is actuated using a Lead Screw (acme) mechanism. The precision stepper motor allows for more control over the downward movement of the system. The stepper motor is controlled with a two-phase stepper motor driver and an encoder for accurate positioning. This also allows for automation of the movement of the drill string. The DC Motor driver is capable of handling analog control signals and is used to provide power to both the top drive and drilling drive motor.

3- Data Acquistion: Sensors discussed are essential to measuring valuable information to increase the efficiency of DREAMS on Mars. They also serve as a key to monitor, in real-time, physical phenomenon of the system which could be used for automation. To collect, process, and interpret this real-time data, a data acquisition system is needed. The signals from the sensors are transferred into a computer employing a Data Acquisition (DAQ) device. Data Acquisition will be carried out using two Arduino Due. The Arduino Due were selected because of they are cost effective, replaceable, reliable and have a high-speed data acquisition rate. DREAMS will be using two Arduino Due which runs at 84 MHz as a discrete stand-alone controller for certain rig's element, e.g. speed controller's reference followers and high-speed data loggers.

4- Power Requirement: DREAMS will be powered by a DC 120V power supply from a wall outlet and the motors and sensors use 12 VDC which is supplied by a switch-mode power AC/D converter rated for a maximum of 10A of current draw. The pump that will be used to extract the water after melting will be a peristaltic pump that will be also powered by 12 VDC. The power system is also limited to the electrical power from the local grid not to exceed 25 hp. Table 1 below summarizes the power requirements for the sensors, motors and actuators. The overall rig system would be a fully automated one. Thus, an open loop flow control was developed to control drilling rate of penetration (ROP). Open loop control will optimize ROP based on drilling formation torque.

Table 1. Power Requirements for the Sensors and Actuators

| Sensors/Actuators | Power Requirement | Specification Name |
|---|---|---|
| **Temperature Sensor** | Voltage: 3v – 5V<br>Max Temp: -40 to +85°C operational range<br>temperature accuracy: +-2°C | GY80-IMU (BMP085) |
| **Inertial Measurement Unit IMU** | Voltage: 3v – 5V | GY80-IMU |
| **Encoder** | Voltage: 5V | AM Mag Encoder |
| **Top drive motor** | Max Power: 363 Watts<br>Voltage: 12 VDC<br>No Load Current: 3.8 AMPs | 775 RedLine Motor |
| **Drilling drive motor** | Max Power: 140 Watts<br>Current: 2.8 Amps<br>Holding torque: 1.95 Nm (276 oz-in) | NEMA 23 Stp-mtr-23079 |
| **Drilling drive motor controller** | Current: 60 Amps<br>Voltage: 12 Volts DC | TB6600 – Stepper motor controller |
| **Peristaltic Pump** | Current: 0.5-1.4Amps<br>Voltage: 12 Volts DC | 12 DC Peristaltic Pump |
| **Heater** | 200-600 watt | Nichrome Wire Heater |

*Challenges*

We faced couple of challenges while designing and building DREAMS

1- Since it was the first time to apply tensegrity structure concepts on drilling rigs, it takes long time and efforts to design, 3D print, fabricate, and test tensegrity parts.
2- Linear and rotary motion of the drilling system within the tensegrity rig structure.
3- We encountered different challenges on how we can insult the casing thermally and electrically from the heating element wires, and also on attaching the casing to a non-rotating part to prevent wire damage while drilling.
4- Due to COVID-19 since March 2020:

a. Texas A&M completely closed the university labs & offices, and we were unable to enter the university premises until July 1st.
   b. Tamu Fischer Engineering Design Center (FEDC) that we rely on for our parts fabrication work to only essential COVID-19 and mission-critical. We were only able to obtain permission to manufacture our parts on June 16$^{th}$.
   c. Difficulties in sourcing components and receiving them in a timely manner -delay in supply chain. Redesigning some components based on parts that could be sourced in a timely manner also added an extra layer of challenges.
   d. Texas A&M imposed a hard restriction on undergraduate students, as they were not allowed to work at all during the summer months within the university premises. Since our team is comprised of a mix of graduate and undergraduate students, we lost a couple of students due to this restriction. Even during full semester, more than half of our team was working remotely.

## *Integration and Test Plan*

The team was able to conduct a couple of test runs:
1- DREAMS successfully drilled four holes in fluffy sand, consolidated sand stone and concrete.
2- Heating, extraction and filtration subsystems were successfully tested together. Our heating element utilized 200 watt power and was able to melt the ice with a rate of 1530 cc/hr. Similarly our filtration unit was capable of cleaning extracted water with a rate equals 750 cc/hr.

## *Project Timeline*

The project timeline is outlined in table 2.

| Month | Activity |
| --- | --- |
| October - January | - Designend Tensegrity rig structure, and water extraction & filtration systems |
| January - March | - Tensegrity rig structure 3D printing<br>- Construction and testing of water filtration system and some electronics |
| March – Mid of June | - Activities stopped due to COVID-19 |
| June 15$^{th}$ – October 15$^{th}$ | - Fabricated tensegrity rig parts in the machine shop<br>- Developed machine learning code<br>- Constructed and tested heating and water extracting systems<br>- Built open loop control system<br>- Conducted four drilling tests on rocks |

## *Safety Plan*

The team has agreed that there should always be safety glasses worn when the drilling rig is in operation. Wearing safety glasses minimizes the chance of debris getting into someone's eye. Ear plugs are also recommended when the drill is in operation to avoid ear damage. Due to a direct result of COVID-19, the team agreed that face masks and gloves should be worn to minimize the spread of respiratory fluids which could contain the virus and only six students or less are allowed to enter the lab at same time based on university rules. The team will also lookout for any water leakage from the water collection system to avoid any member from slipping and causing themselves injury and/or electrocution from the electronics system getting wet.

*Path to Flight*

*Extracting Water on Mars:*

**Tensegrity**

The tensegrity structure concept helps in decreasing the system weight, improve mobility and easier installation in space. The cost of transporting 1 kg of mass to Mars ranges from $4600 to $15,000, hence a lightweight structure will significantly bring the cost of space expeditions for in-situ resource extraction and our experience building the tensegrity structure unraveled key insights. For instance, it is extremely important that the individual components adhere to design measurements during fabrication. Significant time and effort was dedicated in trying to bring the individual components to design shape, post fabrication, as these small deviations in caused major misalignment. Hence, it is strongly advised to assemble the structure on earth to avoid challenges for astronauts.

**Power**

There are no power outlets on Mars, and DREAMS will need a power source to operate by other alternatives. Solar panels are a common solution for many rovers but they have some challenges. The main challenge is Martian dust storms that can last over a month and block out the sunlight partially or completely. This may require very huge solar panel and pose additional burden of weight to provide the necessary power for the system. On the other hand, Multi-Mission Radioisotope Thermoelectric Generators (MMRTG) don't stop working regardless of season or time of day. The issue with MMRTG's is that they have a low power efficiency of 2.8W/kg. In order to save power, we would leverage the sublimation capability. The low pressure on Martian surface, ensures that water stored under the subsurface sublimate on exposure to the atmosphere. Converting this vapor into liquid form for transportation will require further spending of energy; hence, in order to save power, we will transport water in vapor state.

**Weather and Atmosphere**

Temperatures on Mars fall within -140°C to 30°C. DREAMS components should be fabricated of materials capable of withstanding these harsh temperatures variations. Lubricants must have a freezing point lower than -140°C to mitigate bearing seizing. Currently the GY80-IMU, NEMA 23 Stepper motor 23079, TB6600 – Stepper motor controller used in DREAMS contain some plastic parts. Martian temperature is miles below the "glass transition temperature" of plastic resulting in the plastic becoming brittle (Dunn, 2009). The low temperatures also affects the batteries. The best battery system to date are Lithium-ion batteries (LIB), the optimal operating temperature of which varies between 15–35 °C (Ma et al., 2018). The temperature lower than this range affects the performance and may cause irreversible change to the LIBs triggered by the reduction of ionic conductivity and the increase of charge-transfer resistance(Ma et al., 2018). Hence, thermal isolation and low power heaters need to be used to maintain these sensitive components within their operating temperatures. An added benefit of using an MMRTG is that it could help maintain operating temperatures by using excess heat. Due to Mars', thin atmosphere the opposite problem arises. Getting rid of excess heat becomes difficult without convection. Poor convection could also result in having to stop certain operations while a component cools down. Although the winds of Martian storm are not strong enough, the dust particles maybe picked up pose a real threat to the mission. These dust particles could make their way into machinery like gears that could result in serious complications. Countermeasures for this should be taken by sealing everything as well as possible to mitigate the effects of dust.

**Ice Coverage Area**

To ensure the success of the mission the rig must be able to cover more ground. This would increase the likelihood for a more considerable production of water due to the increased areas of exposure. DREAMS future plan involves adopting horizontal drilling to cover more area for water extraction from the same rig slot with less cost and time.

**Water Transport and Storage**

For small quantities of water, it may be feasible to use replaceable and refillable tanks for collection. Colonists, whenever needed, could replace these tanks. For larger quantities, another method for transporting the water is required, because moving individual tanks of water from one place to another is a very time consuming and inefficient process. Alternatively, a system of tubing from the site of the water extraction to the rocket-landing pad would increase the efficiency of the whole process while minimizing the number of necessary Extra Vehicular Activities. However, there is a threat to this method. Due to extreme temperatures, the water in the tubing will freeze and keeping the water in this tubing from freezing will take a considerable amount of energy. It would be more energy efficient to pump the water out as vapor into the tubing system instead of condensing it back down to water first. This would require pumps along the length of the tubing to ensure the vapor does not condense and keeps flowing. Storing the produced water is another challenge, because of the low temperature and pressure on Mars. Water stable state on the surface is solid state, and this makes water pumping out of a storage tank is a challenge. It will also requires a tremendous amount of energy to melt the ice for large quantities of water. If the storage tanks buried below Martian surface, much of the heat energy could be conserved. Digging a large enough hole for a massive water storage tank is challenging even here on Earth. Alternatively, using already existing subsurface cavities such as lava tubes would enable colonists to deploy inflatable storage tanks without having to do much excavation. These underground tanks should be at a reasonable distance away from the colony or rocket landing sites. In the unfortunate event of a collapsed well cavity, much of the surrounding area could be affected. Colony location and water extraction site must be chosen concurrently to guarantee the safety of the colonists. On the other hand, the water storage tanks should not be far. In the case of a leak or some other unforeseen accident, the colonists must be able to react, access the water reserves quickly, and protect such a vital resource.

**Autonomy**

DREAMS cannot operate from the earth, as there would be huge communication delays. Designed system must be able to autonomously dig the well and begin extracting water mostly on its own by a combination of control loop system and machine learning. This can be very challenging due to the severity of conditions but fortunately can be tested on Earth. Once the humans arrive, on Mars they will be able to efficiently operate the system and make repairs wherever necessary.

**Communication**

Communication between DREAMS and earth is essential. Sending a signal such, a long distance takes a lot of energy. This would consume a large portion of the energy reserves depending on the rate of communication. A considerable amount of energy could be saved if Martian satellite is utilized. Martian satellites will be close to the drill rig and save power by sending the signal a shorter distance. Then, these satellites will send these signals back to Earth.

**Extraction and Filtration**

The extraction and filtration subsystems will need some changes for Mars application. The main difference between Earth and Mars operating systems is the ambient environment. Mars is much colder than Earth, as discussed earlier, and has almost no atmosphere. The pressure and temperature differences would cause the ice to sublimate as it is exposed. This presents issues such as refreezing while extracting and losing collected water. Because of this, recirculation as a way of melting the ice and the use of a gravity filter are not viable. In addition, the refreezing the sublimate will consumer excess power. Hence in order to save time and resources, we would transport the water in the form of vapor and have a central system of pressure to both convert into the vapor to liquid form and filter through the gravity filter (as the low gravity on Mars will hinder the process.

**Housing**

The primary structure of support for DREAMS is the tensegrity structure that is light in weight. However, the structure is extremely sensitive to size of its individual components and slight deviation in the sizes during fabrication can pose extreme challenges during assembling the parts. Hence, DREAMS need to be built and assembled on Earth before housing it on a structure as shown in Fig. 7. The structure shall poses the dual capability of providing both anchor and mobility. In order to travel around Mars and its terrain, the housing component will be equipped with six wheels. In addition, the structure will also be provided with six legs that will provide anchor by tucking into the ridges. When an ideal site for excavation is located, the system comes to a halt and the legs will protrude out thereby lifting the structure to provide both leeway to clear the overburden and provide stability while drilling. Accounting for the radiation, the structure will be made of tungsten. Last, the structure will have a door that opens at the bottom to allow the drilling and extraction components to descent.

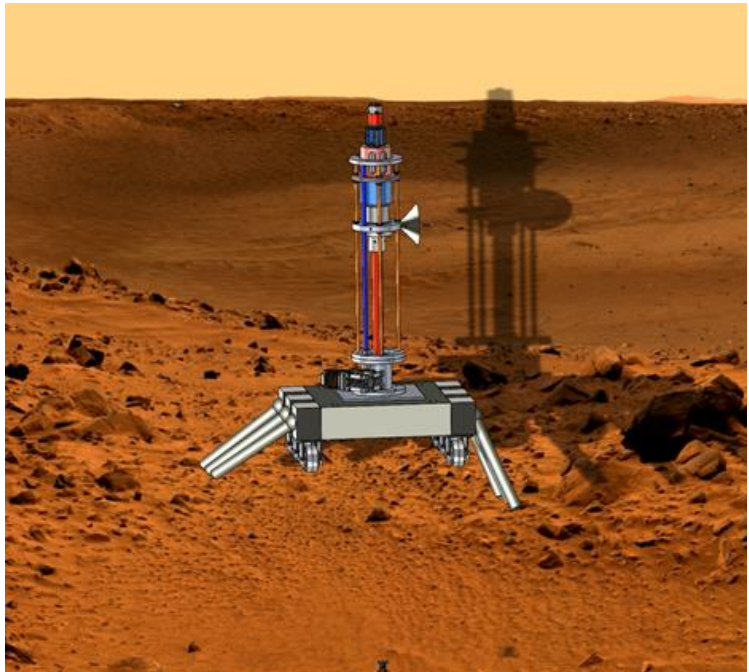

**Fig. 7.** DREAMS housing system concept

**Weather**

The weather on Mars includes frequent dust storms. These storms cause potential harm if DREAMS is not properly protected. Its housing will protect most of the inner subsystems, but some parts are still at risk when the bottom door opens for drilling and extraction. In order to protect the electronics and control systems inside, they will be covered by a shielding made of aluminum. This shielding will protect any sensitive systems from being clogged with dust and any debris.

*Lunar Prospecting*

**Conditions on the Moon**

Gravity and atmosphere are the variables that change significantly between Earth and the Moon; hence, the equipment has to be calibrated accordingly. New mechanisms to hold DREAMS to the surface while drilling will need to be employed due to moon's low gravity (⅙ of earth). In addition, the lack of atmosphere means that there is no protection for the equipment from radiation. The average temperature on the moon undergoes a variation of about 530 degree Fahrenheit and the Moon is much cooler than the earth. Hence, major modifications to DREAMS for lunar prospecting are updates to the drilling and digital core subsystem, material upgrades and a means to traverse lunar terrain and housing.

**Power**

Similar to the discussion in the path to flight section for Mars, lunar adaptation for DREAMS will also use an MMRTG as it will enable digital prospecting on any location, albeit of the position of the Sun.

**Lunar Vacuum**

Most of the greatest engineering challenges on the moon result from its lack of atmosphere. Preventative measures must be taken to avoid cold welding parts together. DREAMS has many moving parts, if any of them get fused together it could result in premature ending to the mission. Materials that are resistant to cold welding should be used wherever possible to reduce the chances of parts seizing up. Very large and efficient radiators must be used to release excess heat. Since there is no convection on the moon it is very difficult to get rid of excess heat. This could greatly affect the operating periods as the operation may need to be stopped to wait for components to cool down. A hard vacuum such as that on the moon causes materials such as plastics to "outgas". When this happens the material greatly degrades and will likely become very brittle and crack and can cause the mission to compromise. For instance, due to an unknown contaminant NASA's Stardust space probe suffered reduced image quality due to an unknown contaminant that had condensed on the CCD sensor of the navigation camera. To avoid these issues only materials, such as metals, that do not outgas must be chosen for any parts that will be exposed to the vacuum. Without an atmosphere, our drill rig will not receive any protection from solar radiation. The chances for a single event upset (SEU) on the moon are much more likely, particularly during the day time. For this reason, the avionics must be much tougher than they would be on Earth. SEU mitigation protocols must be implemented to reduce the chances of an unrecoverable failure on the DREAMS computer. The avionics must also be well isolated to reduce any electrical spikes as a result on surface charges which can also occur the moon.

**Weight on Bit**

To increase the depth of drilling a greater weight on bit is needed. This is a greater concern on the moon than it is on Mars since it has less gravity. Therefore, more weight on bit should be considered while drilling on moon.

**Dust Particles**

Although the moon does not have any winds to pick up dust particles, it is expected that they be kicked up during the drilling process. These dust particles are incredibly fine and pose a serious threat to any moving mechanisms. Substantial sealing must be use to mitigate the possibility for these complications.

**Prospecting on the Moon**

Operating our system on the moon will be faced with similar challenges to the ones encountered on Mars. Temperature on the moon can vary widely between day and night, from -288.4F to 242.6F. With only an infinitesimal amount of air available in its atmosphere when compared to Earth's atmosphere, the atmospheric pressure on the moon is also virtually non-existent. Gravitational acceleration on the moon is approximately ⅙ of that on Earth. Similar changes to the system as suggested in the previous section will also be recommended for lunar operation, with some minor changes. For example, with such an intense temperature swing between day-time and night-time, the casing downhole and other components of the production system must be extremely well-insulated and maintained at an optimal temperature to ensure to lessen the thermal effects on the system's components. Most components that are made of stainless steels in our design, should be replaced by Titanium alloy instead to better withstand the extreme temperature change. Enclosed housing is also required, with an addition of anchoring moors to combat the weak gravitational pull and ensure stability for the system. Additionally, the surface of the moon is known to be exposed to cosmic rays and solar flares, which could have detrimental effects on the electronics of our system. Thus, some forms of radiation shielding (built from materials like Tungsten) that are built into the system housing is necessary. Finally, we see an opportunity to improve the performance of our formation identification and characterization algorithm by training it on a test bed that is composed of simulants representative of the moon formation.

Another crucial step is cooling the drill bit. Due to the constant excavation process, the drill bit will heat up and due to lack of atmosphere, cooling by convection or conduction is not possible. In order to address this issue, we will use water coolant loops, similar to those on the International Space Station, for cooling the system.

## *Budget*

Table 3. Shows the breakout of cost estimates

| Costs | Amount | Grants | Amount |
|---|---|---|---|
| Tensegrity Structure parts | 2812.6 | NASA | 10,0000 |
| Tensegrity Structure Fabrication | 4013.5 | Energy Institute at Texas A&M | 300 |
| Electronics | 2606.2 | Max Vonderbaum Gift Donation | 300 |
| Heating system | 480.8 | Gildin's Class of 1975 DVG Developmental Professorship at Texas A&M. | 400 |
| Water filtration | 993.3 | | |
| Total Expenditures | 10912.1 | Total Budget | 11,000 |
| | Budget Remaining | 87.9 | |

## *Acknowledgment*

- DREAMS team acknowledges financial support from NASA RASC-AL Program, The Energy Institute at Texas A&M, Max Vonderbaum Gift Donation for Drilling Automation Developments and Gildin's Class of 1975 DVG Developmental Professorship at Texas A&M.
- DREAMS Team also acknowledges the support from FEDC and its personnel for timely construction of several parts.
- Special mention goes to John Maldonado form the Petroleum Engineering Department at Texas A&M for tremendous help with building activities and last minute needs.
- The team also thanks Enrqiue Losoya, Narendra Vishnumolakala, Rabih El Helou, Daniyal Ansarid, and Cameron Geresti for his help, valuable comments and suggestions.

## *References*

1- Skelton, Robert E., and Mauricio C. de Oliveira. Tensegrity systems. Vol. 1. New York: Springer, 2009.
2- Nagase, K., and R. E. Skelton. Double-helix tensegrity structures. Aiaa Journal 53.4 (2014): 847-862.
3- Daniel Goldstein, Emilia Kelly, Daniel McGann, Patrick Moore, Andrew Panasuyk, Jacob Rutstein, Ben Zinser, Prospecting Underground Distilling Liquid Extractor, RASC-AL Special Edition Mars Ice Challenge Forum, May 2019.
4- Andrew Adams, Mohsen Alowayed, Roland de Filippi, Ryohei Takahashi, Amy Vanderhout. High Yield Dihydrogen-monoxide Retrieval and Terrain Identification on New worlds. RASC-AL Special Edition Mars Ice Challenge Forum, May 2019.
5- Arjun Krishna, Ann Collins, Nicholas Sorrentino, James Furrer, Dana Roe, Jonathan Bobkov. Drill-based Extraction of Ice-water and Martian Overburden System. RASC-AL Special Edition Mars Ice Challenge Forum, May 2019.
6- Williams, D. R. (2018, September 27). Mars Fact Sheet. Retrieved November 19, 2019, from https://nssdc.gsfc.nasa.gov/planetary/factsheet/marsfact.html.

7- Chaplin, M: (2019, September 9). Water Phase Diagram. Retrieved November 19, 2019, from http://www1.lsbu.ac.uk/water/water_phase_diagram.html.
8- Williams, D. R. (2017, July 3). Moon Fact Sheet. Retrieved November 19, 2019, from https://nssdc.gsfc.nasa.gov/planetary/factsheet/moonfact.html.
9- Barry, P. L. (2005, September 8). Radioactive Moon. Retrieved November 19, 2019, from https://science.nasa.gov/science-news/science-at-nasa/2005/08sep_radioactivemoon.
10- Teale, R. (1965, March). The concept of specific energy in rock drilling. In International Journal of Rock Mechanics and Mining Sciences & Geomechanics Abstracts (Vol. 2, No. 1, pp. 57–73). Pergamon.

*Appendix*
**Appendix A: Tensegrity Rig Component**

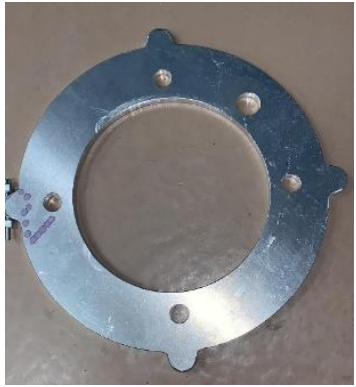

**Fig. 8.** Top and Bottom Ring Design

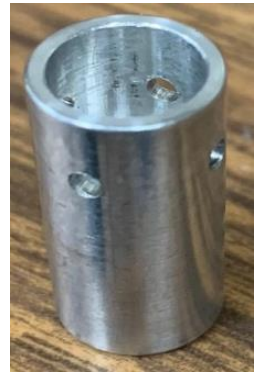

**Fig. 9.** Class 1 joint

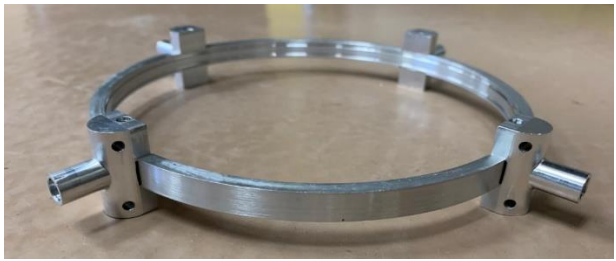

**Fig. 10.** Middle Ring Assembly

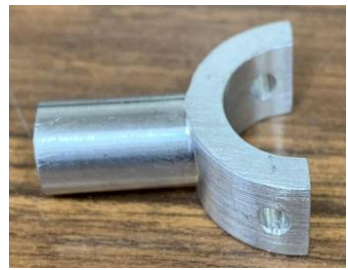

**Fig. 11.** Ring Joint

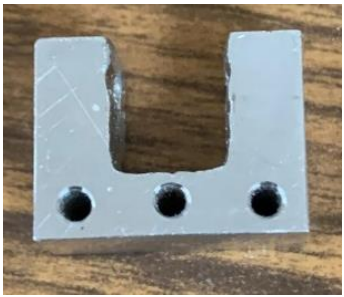

**Fig. 12.** C section for class 2 joint

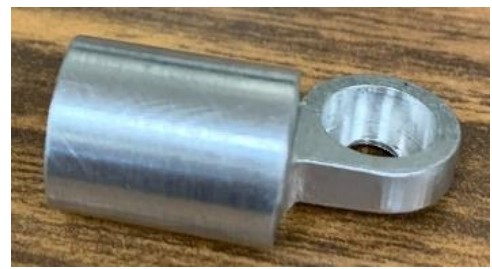

**Fig. 13.** Class 2 joint

## Appendix B: Tensegrity Structure mass

The governing equation of tensegrity statics is given as:

$$NK = W, \quad K = C_S^T \hat{\gamma} C_S - C_B^T \hat{\lambda} C_B, \tag{1}$$

Where N is nodal matrix, where each column represents the coordinates of each node, (where tension members connect to compressive or other tension members), $\gamma$ and $\lambda$ are force density vectors of strings and bars, is a diagonal matrix of the elements of the vector $\gamma$. Cs and Cb are connectivity matrices (with 0, -1, and 1 contained in each column) of strings and bars, and W is external force on the nodes. The atmospheric pressure and the centrifugal forces from the rotation provide all compressive forces needed to stabilize the inflated structure. Yielding is mode of failure for the strings. Using the same material for all the strings, the minimum mass M required for the cable network is:

$$M = \frac{\rho_s}{\sigma_s} \sum_{i=1}^{\alpha} \gamma_i \|s_i\|^2 + \sum_{j=1}^{\beta} max\left(\frac{\rho_b}{\sigma_b} \lambda_j \|b_j\|^2, 2\rho_b \lambda_j^{\frac{1}{2}} \left(\frac{\|b_j\|^5}{\pi E_b}\right)^{\frac{1}{2}}\right), \tag{2}$$

Where \rho_s and \sigma_s are density and yield strength of strings and where \rho_b, \sigma_b, and E_b are density, yield strength, and Young's Modulus of bars. Material used for strings: Spectra (Ultra High Molecular Weight Polyethylene, UHMWPE). Aluminum: E_b = 60e09 Pa, rho_b = 2700 kg/m3, sigma_b = 110e06 Pa. UHMWPE Density: E_s = 120e09 Pa, rho_s = 970 kg/m3, sigma_s = 2.7e09 Pa.

### Dynamics of the structure

It is also important to know the dynamic behavior of the structure. A compact matrix form for the full system dynamics including string masses can be written as [7]:

$$\ddot{N} M_s + N K_s = W + \Omega P^T,$$
$$K_s = \left[ C_s^T \hat{\gamma} C_{sb} - C_{nb}^T C_b^T \hat{\lambda} C_b \quad C_s^T \hat{\gamma} C_{ss} \right],$$
$$\hat{\lambda} = -\hat{I} \hat{l}^{-2} [\dot{B}^T \dot{B}] - \frac{1}{2} \hat{l}^{-2} [B^T (W + \Omega P^T - S \hat{\gamma} C_s) C_{bn}^T C_b^T]. \tag{3}$$

where Cnb, Cb, Csb, Csb, Css represent different connectivity matrices for bar to string, string to string and bar to bar nodes, is the Lagrange multiplier, P is the constraints matrix.